\documentclass[12pt]{article}

\usepackage{psfrag}
\usepackage{color}
\usepackage{algorithm}
\usepackage{algorithmic}
\usepackage{amsfonts}
\usepackage{amsmath}
\usepackage{mathrsfs}
\usepackage{amssymb}
\usepackage{wrapfig}
\usepackage{graphicx}
\usepackage{psfrag,subfigure} 
\usepackage{amsthm}

\newcommand{\bbeta}{\boldsymbol{\beta}}

\newcommand{\bA}{\mathbf{A}}
\newcommand{\bW}{\mathbf{W}}
\newcommand{\bH}{\mathbf{H}}

\newcommand{\bX}{\mathbf{X}}

\newcommand{\bh}{\mathbf{h}}

\newcommand{\by}{\mathbf{y}}
\newcommand{\bx}{\mathbf{x}}

\newcommand{\bw}{\mathbf{w}}
\newcommand{\argmin}{\operatornamewithlimits{argmin}}

\newtheorem{theorem}{Theorem}

\definecolor{violet}{RGB}{143, 0, 255}

\newcommand{\xxx}{STRADS}

\usepackage{JASA_manu}
\usepackage{hyperref}

\usepackage{color}
\def\bSig\mathbf{\Sigma}

\begin{document}

\title{Structure-Aware Dynamic Scheduler for Parallel Machine Learning}
\author{Seunghak Lee,  
Jin Kyu Kim,
Qirong Ho, \\
Garth A. Gibson, 
Eric P. Xing$^{*}$ \\ 
School of Computer Science \\ 
Carnegie Mellon University, Pittsburgh, PA, U.S.A. \\ 
$^{*}$email: \texttt{epxing@cs.cmu.edu} }

\maketitle

\begin{center}
\textbf{Abstract}
\end{center}
Training large machine learning (ML) models with many variables or parameters can take a long time 
if one employs sequential procedures even with stochastic updates. 
A natural solution is to turn to distributed computing on a cluster; 
however, naive, unstructured parallelization of ML algorithms does 
not usually lead to a proportional speedup and can even result in divergence, 
because dependencies between model elements can attenuate the computational 
gains from parallelization and compromise correctness of inference. 
Recent efforts toward this issue have benefited from exploiting 
the static, {\it a priori} block structures residing in ML algorithms. 
In this paper, we take this path further by exploring the dynamic block structures 
and workloads therein present during ML program execution, which offers new opportunities 
for improving convergence, correctness, and load balancing in distributed ML. 
We propose and showcase a general-purpose scheduler, STRADS, for 
coordinating distributed updates in ML algorithms, which harnesses the aforementioned opportunities in a systematic way.
We provide theoretical guarantees for our scheduler, and demonstrate its efficacy versus static block structures on Lasso and Matrix Factorization.

\vspace*{.3in}

\section{Introduction}

Sensory techniques and digital storage media have improved at a breakneck pace, leading to massive collections of data. The resultant so-called Big Data problems have been a common focus in recent enthusiasms toward scalable machine learning, and numerous algorithmic and system solutions have been proposed to alleviate the time-bottleneck due to Big Data by exploring various heuristic or principled strategies for {\it data parallelism}~\cite{dean2008mapreduce,low2012distributed,malewicz2010pregel,zaharia2010spark}.

However, another important aspect of Big ML is what we refer to as {\it Big Model} problems, in which models with millions if not billions of variables and/or parameters (e.g., as one would see in a deep network or a large-scale topic model) must be estimated from big (or even modestly-sized) data; such Big Model problems seem to have received relatively less attention from the community.
In this paper, we investigate how to facilitate effective and sound parallelization of inference over a large number of variables and/or parameters in such models, an issue we call {\it model parallelism}. 
Model-parallel inference is necessary for many ultra-high dimensional problems that have recently emerged in modern applications. For example,  
in genetics and personalized medicine,
the number of model variables (e.g., candidate genetic variations) can easily exceed millions; and 
in e-commerce applications such as personalized ads recommendation, the 
so-called ``interest genome'' derived for every person from their multi-media social trace is also very high-dimensional.
These high-dimensional problems must be solved quickly to be practically useful for patients or consumers, therefore sequential computation even on a powerful, high-end single machine is usually not an option, and distributing the computation over large number of processors in a cluster becomes a natural choice.
It is important to note that model-parallelism is not the same as data-parallelism, and poses very different challenges --- 
model-parallelism requires model variables to be partitioned for parallel updates with tight synchronization, 
whereas data-parallelism involves computation on (usually) independent data subsets. 
Our focus in this paper is to systematically study the algorithm, system, and theory issues that support model-parallelism.

A major challenge to model-parallelism is that many existing algorithms for ML are derived with the 
assumption of {\it sequential iteration}
over variables ---
for example, optimization algorithms for Lasso \cite{tibshirani1996regression}, 
matrix factorization \cite{yu2012scalable}, sparse coding \cite{gregor2010learning}, and
support vector machines \cite{hsieh2008dual}; MCMC algorithms for
topic models \cite{griffiths2004finding}, Bayesian nonparametric models \cite{ghosh2003bayesian},
and direct posterior regularization models \cite{ganchev2010posterior}.
However, the convergence rates and correctness guarantees for these algorithms do not always extend to
the parallel execution over model variables. 
In other words, naive model-parallelization can slow down the convergence rate
or even lead to failure of ML algorithms~\cite{bradley2011parallel}.

In this paper, we focus on the problem of how to parallelize ML algorithms 
over different model variables.  
Recent efforts toward parallel ML over model variables can be divided into two approaches:
(1) unstructured distributed ML, and
(2) structured distributed ML.
The first approach includes algorithms that 
select model variables uniformly at random for parallel execution 
\cite{bradley2011parallel};
the second approach uses the problem structure
to select which variables to update in parallel,
thus speeding up per-iteration convergence rates \cite{scherrer2012feature}, boosting iteration frequency by improving
distributed system performance 
(e.g. minimizing network communications or disk I/O) 
\cite{kang2009pegasus,kyrola2012graphchi}, 
and guaranteeing algorithm correctness \cite{scherrer2012feature}.
While structured distributed ML has benefits over unstructured approaches,
there is an additional cost to finding such structures. In this paper, 
we adopt the approach of structured distributed ML, 
but in a way that departs significantly from conventional strategies, as inspired by the following insights on how structures in an ML program can be explored and exploited.

\paragraph{Static Block Structures:}
Static block structures, which are often assumed to be intrinsic to a model and discovered before algorithm start
and held fixed during execution,
have been widely used for efficient parallerization of ML algorithms.
Examples include block Gibbs sampling \cite{lee2009convolutional}, 
structured mean field approximations \cite{jaakkola200110},
graph-partitioning for parallel executions in GraphLab \cite{low2012distributed}, 
parallel coordinate descent for matrix factorization \cite{yu2012scalable}, 
and block-greedy coordinate descent algorithm \cite{scherrer2012feature}. 
The key insight of this approach is
that, if decoupled blocks of variables are updated in parallel, 
then both inconsistencies due to parallel variable updates as well as
communications between different blocks are minimized, 
resulting in improved ML algorithm convergence rates and guaranteed correctness.
However, static block structures must be discovered prior to starting
an ML algorithm, resulting in a large, unavoidable runtime cost.
Furthermore, static block structures fail to capture the dynamic, changing aspects of 
ML algorithms driven by data (such as how variables and parameters change throughout execution), and
thus cannot obtain a holistic view of an ML problem's structure.

\paragraph{Dynamic Block Structures:}
In reality, model block structures are not completely static,  
but can dynamically change at runtime according to the values of parameters and variables due to data-driven updates.
Notably, transient block structures can arise due to recently
updated parameters and variables.
Taking $\ell_1$-regularized regression as an example, let us consider 
parallel updates of $\{\beta_j, \beta_k\}$ at the $t$-th iteration.
If $\beta_k$ stays zero at the $(t-1)$-th and $t$-th iteration, then
$\beta_k^{(t)}$ does not affect the update of $\beta_j^{(t)}$ --- even 
when the correlation (i.e. dependency) between $\bx_j$ and $\bx_k$ is large. 
Such dynamic, runtime structure discovery is critical 
for distributed ML algorithms,
because static block structure, while useful, relies on
finding separable blocks from the input data and the model's a priori topology ---
a challenging task on many real datasets \cite{witten2005data}.  
Furthermore, dynamic block structures have computational advantages 
over their static counterparts --- because dynamic structure is
discovered online during algorithm execution, its cost can be amortized over
multiple processors working in the background (as opposed to the lump sum
cost of static structure discovery).

\paragraph{Projected-Progress on Dynamic Block Structures:}
While dynamic block structures ensure correct parallel execution,
they do not directly expedite or improve ML algorithm convergence rates --- for that, it
is necessary to account for
(1) each block's projected progress or importance (e.g. expected objective value improvement upon updates)
and (2) the actual workload of each block (e.g. number or magnitude of variables to be updated),
when dispatching blocks of variables to parallel workers. Continuing the $\ell_1$-regularized regression example, if we prioritize the $\beta_j$ that are changing the most with each update, we will speed up the decrease in the loss function per variable update. Moreover, by ensuring every worker gets a similar number of variables to update, we perform {\it load-balancing}, thus preventing situations where workers with fewer variables end up sitting idle.

\begin{figure}[t] 
\centering
\includegraphics[width=0.6\textwidth]{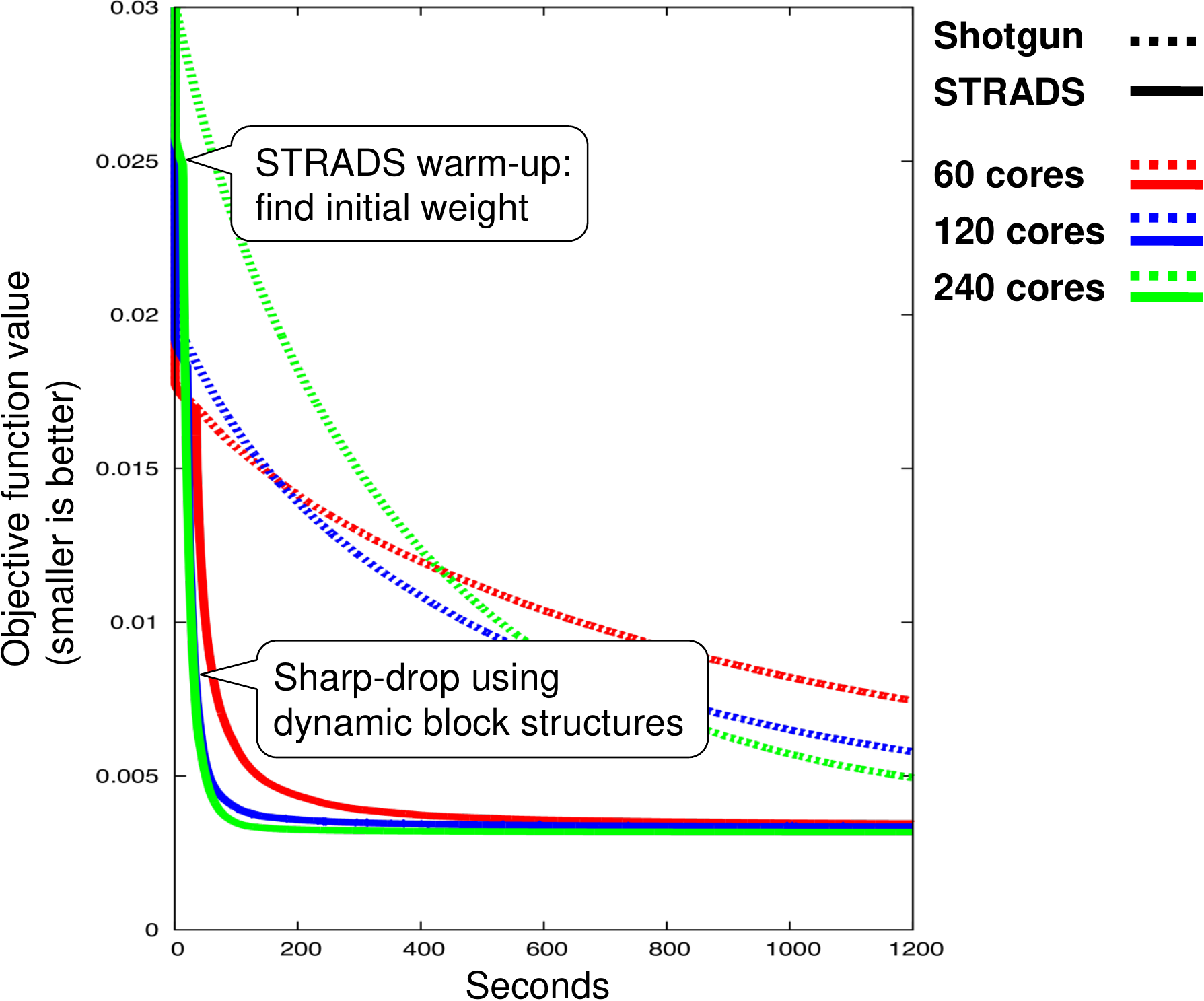}
\caption{
Convergence rates of two different approaches for parallel Lasso:
dynamic block structures (STRADS), 
and no structures (Shotgun \cite{bradley2011parallel}) (We used 
Alzheimer's disease dataset \cite{sagebionetworks} 
with $\lambda = 5 \times 10^{-4}$).}  
\label{fig:intro_exp}
\end{figure}

In this paper, we seek to explore {\it Structure-Aware Parallelism} (SAP) inspired by the above insights, at both the algorithmic front-end and system back-end levels. Accordingly, we present a new model-parallel ML strategy called \xxx{}, or {\it STRucture-Aware Dynamic Scheduler}. Figure~\ref{fig:intro_exp} showcases the key advantage resulting from this new strategy: under a dynamic block-based parallel approach, the convergence of an ML algorithm can escape from the slow-progressing trajectory characteristic of a static block-based parallelism, thus arriving at a better solution much more quickly. This is made possible by the dynamic approach's ability to adapt to changing structure and execution status, as an ML program (in this case, parallel Lasso) progresses.

\begin{figure}[t] 
\centering
\includegraphics[width=0.8\textwidth]{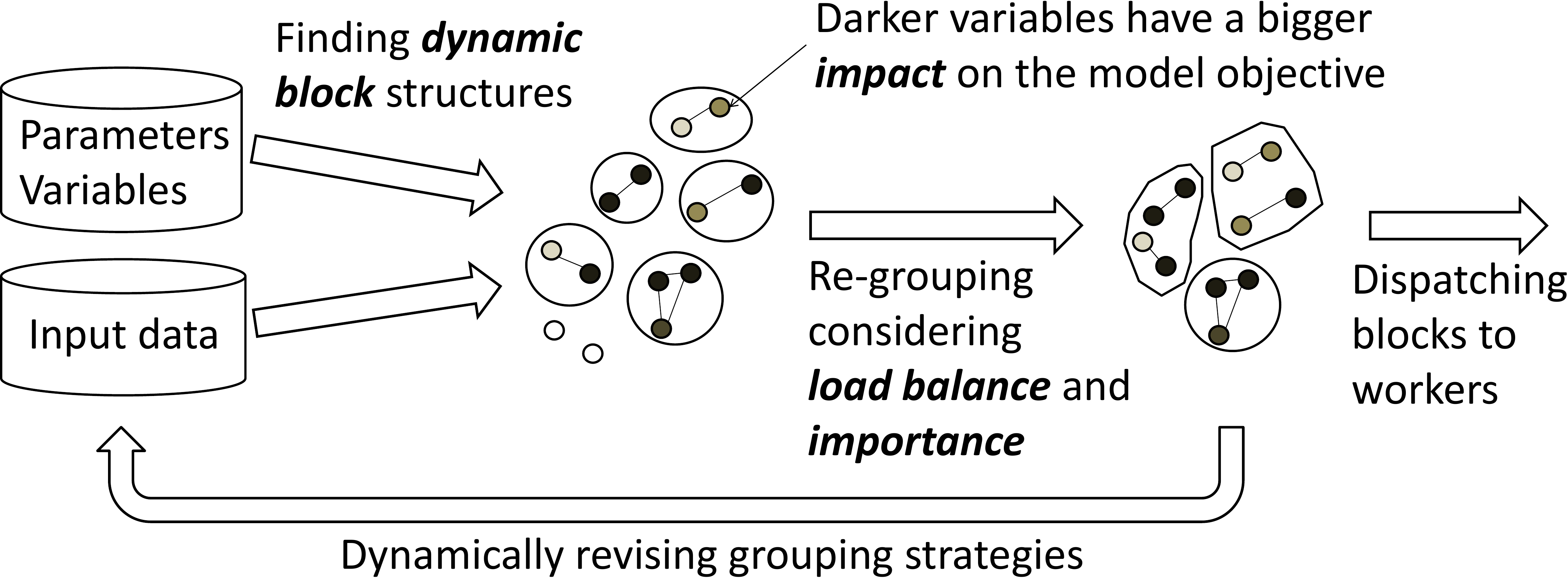}
\caption{
Concept diagram for the SAP scheduling model underlying \xxx{}, explaining how blocks of model variables are selected, grouped and dispatched to workers.}
\label{fig:strads_concept}
\end{figure}

More precisely, \xxx{} is a statistically motivated scheduler that executes distributed ML algorithms correctly and with high-convergence rates by jointly considering dynamic block structures, load balancing, and algorithmic progress made by updates on variable blocks. Figure \ref{fig:strads_concept} outlines the basic rationales behind \xxx{}, while Figure~\ref{fig:scheduler} sketches out the system architecture. We apply STRADS to two example applications: parallel Lasso and parallel matrix factorization (MF), using coordinate descent algorithms (we expect \xxx{} applies to other algorithms on additional ML programs, which we will subject to future case studies). In the Lasso example, we showcase the benefits of using dynamic block structures based on the runtime values of coefficients; and in the MF example, we demonstrate the advantages of load balancing for parallel execution. Furthermore, we provide a theoretical analysis that proves our scheduling scheme for Lasso is approximately optimal. Our experiments show that for Lasso and MF, STRADS yields faster convergence than unstructured and
static block-based approaches, as well as better final objective function values (for Lasso).

\paragraph{Notation} 
We denote a matrix with $N$ samples and 
$J$ variables as $\bX \in \mathbb{R}^{N\times{}J}$, 
and a vector with $N$ samples as $\bx \in \mathbb{R}^{N\times{}1}$.
We represent a column index by subscript, and a row index by superscript. 
We denote iteration index by parenthesized superscript, 
matrices by bold-faced 
uppercase letters, and vectors by bold-faced lowercase letters.
We also denote the ``dependency strength" (e.g. correlation)
between the $j$-th and $k$-th variables by $d(\bx_j,\bx_k)$, and
a set of $M$ blocks of variables by $\mathcal{B} = \{B_1,\ldots,B_M\}$.

\section{Structure-Aware Parallelism (SAP) for Dynamic Block Scheduling}

We begin with an outline of the scheduling model upon which our proposed approach is built, which we call
Structure-Aware Parallelism (SAP).
Suppose there are $J$ model variables, and we have $P$ parallel workers to update them.
SAP iterates over four steps: 

{\small
\begin{enumerate}
\item Draw a set $\mathcal{U}^{(t)}$ of $P'(>P)$ variables to update at iteration $t$,
from an ``importance" distribution $p(j)$. 
The idea is to choose variables that give high expected improvement
in the loss function.\\[-0.4cm]
\item From the set $\mathcal{U}^{(t)}$,
find a set of variable blocks $\mathcal{B}_1$
such that $D(B_j,B_k) < \rho$, where $B_j, B_k \in \mathcal{B}_1$ and
$\rho$ is a user-defined parameter.
Here, $D(B_j,B_k)=\max{|d(\bx_l,\bx_m)|}$, where $\bx_l \in B_j, \bx_m \in B_k$,
and $d(\bx_l,\bx_m)$ is a user-defined measure of coupling between $\bx_l$ and $\bx_m$ 
(e.g. correlation or partial correlation).\\[-0.4cm]
\item Merge blocks of variables in $\mathcal{B}_1$ until every block has a similar
workload, thus achieving load balance. We denote
the set of regrouped blocks by $\mathcal{B}_2$, 
and we dispatch the $P$ blocks of variables in $\mathcal{B}_2$ 
to $P$ workers. \\[-0.4cm]
\item When the $P$ workers have finished and returned their updated blocks to SAP,
SAP updates the importance distribution $p(j)$ and the
dependency function $d(\bx_j,\bx_k)$ according to the updated blocks.
SAP then iterates steps (1)-(4) until the ML algorithm converges.\\[-0.6cm]
\end{enumerate}
}

The first step maximizes convergence rates, by selecting variables
that will contribute the most to loss function improvement. This is carried out by sampling variables from the distribution $p(j)$, which assigns higher probability to variables whose recent
updates have had higher impact on the loss function.
Note that $p(j)$ changes across iterations, because the importance
of each variable changes as the algorithm progresses ---
this is a key difference between dynamic and static block structures.

The second step ensures correctness of an ML algorithm,
by decoupling only those blocks of variables with little to no interdependency.
It is well-known that simultaneous updates to strongly coupled variables
can cause interference; this not
only slows down convergence but can even lead to
algorithm failure (e.g. divergence)~\cite{bradley2011parallel}.
By organizing variables into nearly-independent blocks, we can 
control the degree of interference, thus guaranteeing correctness.

The third step performs load balancing. 
Because blocks can greatly vary in size, situations can arise where most workers
end up waiting for the worker with the biggest block to finish --- the
``curse of the last reducer"~\cite{suri2011counting}. 
We address this problem by merging blocks until all remaining blocks contain similar workloads.

The fourth step is a ``progress monitoring" step, in which SAP estimates the progress
each variable contributes to algorithm convergence. Depending on the ML algorithm being
run, the definition of progress can vary: examples include the magnitude of change in each variable,
or the change in residuals due to variable updates. SAP then uses this information to
update $p(j)$ (e.g. by increasing the probability of faster-changing variables) and
$d(\bx_j,\bx_k)$ (e.g. by removing dependencies between variables that have reached zero).

We note that SAP is only one scheme for dynamic block scheduling,
and other designs are possible. The key advantage of our
design is computational efficiency:
step 1 minimizes the computational cost of scheduling by 
reducing the set of variables from which we must find block structures ---
essentially a bootstrap-based approach to structure discovery.
This is important because the scheduler must be able to find block
structures faster than workers consume them (i.e. the scheduler must
not be a bottleneck).

We now showcase how coordinate descent algorithms for two popular
ML models, $\ell_1$-regularized regression and Matrix Factorization, can be cast into parallel versions  via SAP.

\subsection{Case 1: $\ell_1$-regularized Regression}
\label{sec:lasso}

\begin{algorithm}
\caption{Parallel CD for Lasso, using SAP}
\begin{algorithmic}
\STATE Choose $P$, $\eta$ (a small positive constant), and $\rho$ 
\STATE Set $t_j=2$ for all $j=1, \ldots, J$, where $t_j$ represents the 
iteration-counter for the $j$-th coefficient
\STATE Set $\beta_j^{(t_j-2)} = C, \beta_j^{(t_j-1)} = 0$
for all $j=1, \ldots, J$, where $C$ is a very large positive constant 
\WHILE{\mbox{not converged}}
\STATE Draw a set $\mathcal{U}$ of $P'$ coefficients, 
from a distribution $p(j) \propto |\beta_j^{(t_{j}-2)}-\beta_j^{(t_{j}-1)}| + \eta$, 
where $P'$ is a constant larger than $P$
\STATE Choose a set $\mathcal{B}_1$ of $\leq P$ coefficients (i.e. one-variable blocks) from $\mathcal{U}$, such that $|\bx_j^T \bx_k| \leq \rho$ for all 
$\bx_j, \bx_k \in \mathcal{B}_1$, where $\bx_j$ is the covariate corresponding to $\beta_j$
\STATE {\bf In parallel} on $P$ workers
\STATE \quad Get assigned coefficient $j$ from $\mathcal{B}_1$
\STATE \quad Update $\beta_j^{(t_j)}$ using update rule \eqref{eq:update_rule}
\STATE \quad $t_j \leftarrow t_j + 1$;
\ENDWHILE
\end{algorithmic}
\label{alg:our_algorithm}
\end{algorithm}

The $\ell_1$-regularized regression (a.k.a Lasso) \cite{tibshirani1996regression} 
is used to discover a small subset of
features or dimensions that 
are relevant to an output $\by$. 
$\ell_1$-regularized regression takes the form of an optimization program: 
\begin{align}
\min_{\bbeta} L(\bX, \by, \bbeta) + \lambda \sum_{j} |\beta_j|
\label{eq:lasso}
\end{align}
where $\lambda$ denotes the regularization parameter that can be tuned, and $L(\cdot)$ is a non-negative convex loss function such 
as squared-loss or logistic-loss; 
we assume that $\bX$ and $\by$ are standardized and consider \eqref{eq:lasso} without an intercept.
Throughout this paper, for simplicity but without loss of generality, we let  
$L(\bX,\by,\bbeta) = \frac{1}{2}\left\| \by - \bX \bbeta \right\|_2^2$.
However, 
it is straightforward to use other loss functions such as logistic-loss using the same approach shown in \cite{bradley2011parallel}.

By taking gradient of \eqref{eq:lasso}, we obtain the coordinate descent (CD)
algorithm \cite{friedman2007pathwise} update rule for $\beta_j$:
\begin{align}
\beta_j^{(t)} \leftarrow S(
\bx_j^T\by 
-\sum_{k \neq j} \bx_j^T \bx_k \beta_k^{(t-1)},\lambda),
\label{eq:update_rule}
\end{align}
where $S(\cdot,\lambda)$ is a soft-thresholding operator \cite{friedman2007pathwise}.

SAP schedules parallel CD updates on the Lasso optimization program \eqref{eq:lasso},
according to the four steps:\\[-0.6cm]

\begin{itemize}
	\item {\bf Step 1}: 
We use probability $p(j) \propto \delta \beta_j$, where
$\delta \beta_j = |\beta_j^{(t)}-\beta_j^{(t-1)}|$,
where $\beta_j^{(t)}$ represents $\beta_j$
at the $t$-th iteration.
Intuitively, convergence is improved when we
update variables (coefficients) that change more rapidly per iteration, and thus
we prioritize variables based on their value change. 
In Section \ref{sec:theory}, we provide a theoretical justification for use of $p(j)$.\\[-0.4cm]

	\item {\bf Step 2}: We define dependency 
$d(\bx_l,\bx_m)$ for the parallel updates of \eqref{eq:update_rule}. In this case, 
$d(\bx_l,\bx_m)=|\bx_l^T \bx_m|$, i.e., correlation between 
$l$-th and $m$-th covariates (note that we standardized $\bX$). 
If the $j$-th and $k$-th covariates, $\bx_j$ and $\bx_k$, are highly correlated,
then updating $\{\beta_j,\beta_k\}$ in parallel will cause an interference effect
that may dramatically attenuate improvement 
in the objective function \cite{bradley2011parallel}.
SAP ensures that variables $\beta_j$ are grouped into blocks
such that variables in different blocks have nearly independent 
covariates $\bx_j$ --- thus keeping intereference effects to a minimum.\\[-0.4cm]

	\item {\bf Step 3}: For parallel Lasso, we fixe the size of blocks to one for application-specific
reasons. It turns out that it is non-trivial to choose 
an appropriate size of blocks considering both 
load balance and quality of updates (i.e., decrease of objective value).
Thus, choosing an appropriate size of blocks at runtime
is left for future work. \\[-0.4cm]

    \item {\bf Step 4}: After collecting the updated variables $\beta_j^{(t)}$
from workers, SAP uses them to update $p(j)$ from Step 1. \\[-0.6cm]
\end{itemize}

\subsection{Case 2: Matrix Factorization}

MF is often used for collaborative filtering, where the goal is to
predict a user's unknown preferences, given his/her known 
preferences and the preferences of others. The input data is modeled 
as an incomplete matrix
$\bA \in \mathbb{R}^{N\times{}M}$, where $N$ is the number of users, and $M$ is the number of items/preferences.
The idea is to discover smaller rank-$k$ matrices 
$\bW \in \mathbb{R}^{N\times{}K}$ and $\bH\in \mathbb{R}^{K\times{}M}$
such that $\bW \bH \approx \bA$. Thus, the product $\bW\bH$ can be used
to predict the missing entries (user preferences).
Formally, let $\Omega$ be the set of indices of observed entries in $\bA$, 
$\Omega^i$ be the set of observed column indices in the $i$-th row of $\bA$, and
$\Omega_j$ be the set of observed row indices in the $j$-th column of $\bA$.
Then, the MF problem is defined as an optimization program
\begin{equation}
\min_{\bW,\bH} \sum_{(i,j) \in \Omega} (a_{j}^i - \bw^i\bh_j)^2 + \lambda ( \left\| \bW \right\|_F^2 + \left\| \bH \right\|_F^2).
\label{eq:mf}
\end{equation}
This optimization is solved via parallel CD, with the following update rules for $\bW$ and $\bH$:
\begin{align}
w_t^i & \leftarrow \frac{\sum_{j \in \Omega^i}(r_j^i+w_t^i h_j^t)h_j^t}{\lambda + \sum_{j\in \Omega^i} (h_j^t)^2},\\
h_j^t & \leftarrow \frac{\sum_{i \in \Omega_j}(r_j^i+w_t^i h_j^t)w_t^i}{\lambda + \sum_{i\in \Omega_j} (w_t^i)^2},
\end{align}
where $r_j^i = a_j^i - \bw^i \bh_j$ for all $(i,j) \in \Omega$.

To solve the MF problem, SAP iterates through each rank $t\in\{1\dots K\}$, parallelizing the updates
$\{w_t^i\}$ over blocks of rows $i$ in $\bA$, 
and parallelizing the updates
$\{h_j^t\}$ over blocks of columns $j$ in $\bA$. Specifically:\\[-0.6cm]

\begin{itemize}
\item {\bf Step 1}: 
For MF, prioritizing variables within a full
column $\bW$ or row $\bH$ results in minimal benefit,
hence we use a uniform distribution for $p(j)$.\\[-0.4cm]
\item {\bf Step 2}: 
In MF, each coefficient can be independently updated without interference.
Thus, $d(\bx_l, \bx_m)=0$, and any coefficients can be grouped together.\\[-0.4cm]
\item {\bf Step 3}: 
Because the observed matrix entries often follow a power-law distribution,
we perform load balancing by grouping rows and columns into larger blocks,
such that the nonzero entries of $\bA$ are equally distributed.\\[-0.4cm]
\item {\bf Step 4}:
Since $p(j)$ and $(d\bx_l,\bx_m)$ are constant functions, no modification is required.\\[-0.6cm]
\end{itemize}

\section{\xxx{}: an Efficient, Distributed Implementation of SAP}

\begin{figure}[t] 
\centering
\includegraphics[width=0.55\textwidth]{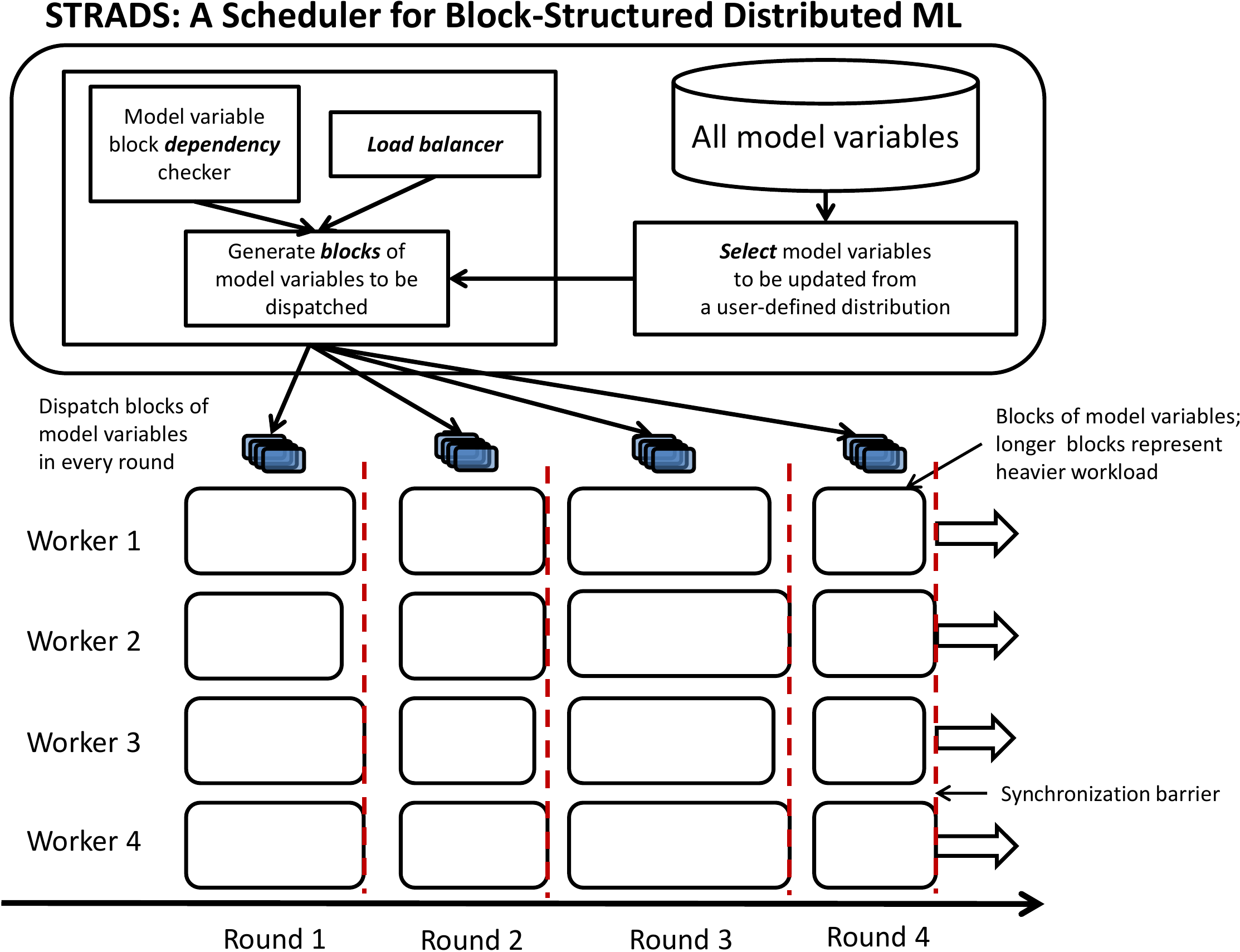}
\caption{ 
{\bf A high-level view of the \xxx{} architecture.}
\xxx{} begins by selecting model variables from an importance distribution $p(j)$ (to improve convergence rate),
and then groups these variables into blocks according to a dependency function $d(\bx_j,\bx_j)$ ---
the idea being to avoid scheduling highly-dependent variables in parallel on different blocks
(to maintain correctness).
\xxx{} then merges these blocks into larger blocks (for load balancing), and then dispatches
these load-balanced blocks to workers. The workers then report their updated blocks back
to \xxx{} which uses them to update the importance distribution $p(j)$ and dependency function $d(\bx_j,\bx_j)$.
This process constitutes one iteration, which is repeat with a new set of blocks (i.e. dynamic structure).
Our implementation of \xxx{} is fully distributed over multiple machines.}
\label{fig:scheduler}
\end{figure}

Now we describe \xxx{} (Figure \ref{fig:scheduler}), a distributed implementation of the SAP scheduling model,
which can use any number of machines to provide scheduling for an arbitrary degree of pallellism.
Having a distributed implementation ensures that STRADS will scale to meet
the computation and memory demands of finding dynamic block structure on
extremely large models and input data.
The key ideas behind STRADS are (1) each scheduler thread is
responsible for scheduling its own disjoint set of variables (and only those variables),
and (2) the scheduler threads take turns to send blocks to the worker clients.

\paragraph{Implementation Overview}
Suppose the user invokes $S$ \xxx{} threads (which can be on different machines)
to solve an ML model with $J$ variables. \xxx{} proceeds as follows:
First, each thread $s$ is randomly assigned $J/S$ variables (with no overlaps) before
the algorithm starts; these assignments remain fixed throughout. Next,
all threads execute the four SAP steps --- (1) select $P'$ variables from $p_s(j)$, where $p_s(j)$ is the importance distribution over the $J/S$ variables assigned to thread $s$, (2) use those $P'$ variables to form the set of dynamic variable blocks $\mathcal{B}_1$ according to $d(\bx_j,\bx_k)$, (3) merge blocks to get a new set of blocks $\mathcal{B}_2$ that are load-balanced, and distribute them to workers, and (4) receive the updated blocks from workers, and update $p_s(j)$, $d(\bx_j,\bx_k)$.
The \xxx{} scheduler threads take turns to dispatch $\mathcal{B}_2$ to the $P$ workers:
thread 1 dispatches first, then thread 2, and so on until thread $S$,
before returning to thread 1.

In our experiments, we assume the entire input data
is available to every machine, though we note that it can just
as easily be stored in a distributed key-value store or parameter server.
Each scheduler thread maintains and stores only the $J/S$ variables assigned to it.
We implement \xxx{} in C++, using the Boost libraries and the
0MQ 3.2.4 library \cite{hintjens2013zeromq} for inter-machine network communications.

\paragraph{Programming Interface}
Per the SAP model, \xxx{} requires users to define model-specific functions
$p(j)$ and $d(\bx_j,\bx_j)$, via the following interface:
\begin{itemize}
\item
{\tt define\_sampling(p)}, where {\tt p} is a function object such that {\tt p(j)} returns the probability of variable {\tt j}. \xxx{} also provides {\tt p} with an interface to access the input data, as well as the model variables (on the current \xxx{} thread); we shall not go into the details for space reasons.
\item
{\tt define\_dependency(d)}, where {\tt d} is a function object such that {\tt d(j,k)} returns the dependency between variables {\tt j} and {\tt k}. Like {\tt p}, {\tt d} has access to the model variables and input data through \xxx{}.
\end{itemize}

\paragraph{Properties of \xxx{}}

From a distributed systems perspective, the round-robin design of \xxx{} carries the following benefits: one, it makes effective use of distributed cluster memory --- every scheduler thread only needs to store the state of the $J/S$ variables assigned to it. Two, the scheduler threads require almost no communication between each other; they just need to coordinate taking turns to serve workers. Three, the round-robin arrangement allows each scheduler thread more time to prepare $\mathcal{B}_2$ for dispatch --- if there are $S$ threads, then each thread has $S$-fold more time. This prevents situations where workers have to wait for the schduler, and is essentially a form of hiding computational latency.

\xxx{} is essentially a bootstrap of the SAP model. Even though the importance distribution $p(j)$ is now split into $S$ distributions $p_s(j)$, since $J \gg S$ for Big Model problems, each $p_s(j)$ will be approximately similar in shape to the original $p(j)$. Furthermore, \xxx{} preserves algorithm correctness: because blocks from different scheduler threads will be updated at different iterations, there is no need to cross-check depedencies for blocks between threads. Load balancing is also unaffected, provided that $J/S$ is sufficiently large (so that enough blocks are produced). Thus, \xxx{} is a close, bootstrapped approximation to SAP scheduling for Big Models with 
a large number of variables and/or parameters.

\section{Theoretical Analysis of Parallel CD Under SAP}
\label{sec:theory}

The SAP model specifies a general-purpose dynamic block scheduler for distributed ML algorithms; given a specific ML algorithm, the user must input appropriate definitions for $p(j)$ (important variable subsampling) and $d(\bx_j,\bx_k)$ (dependency checking).
To provide theoretical analysis of parallel CD under SAP, let us consider the definitions
for Lasso regression in Section \ref{sec:lasso} --- under them,
we will show that SAP approximately obtains the
optimal Lasso convergence rate for $P$ worker threads.
We formally re-state those definitions:
\begin{enumerate}
	\item Select a subset of variables in the $t$-th iteration:  
	choose $P'>P$ Lasso coefficients (variables), i.e., $\mathcal{P}_t = \{v_j\}_{j=1}^{P'}$, 
	where $v_j \in \{1, \ldots, J\}$ are selected from the distribution
	$p(j) \propto \delta \beta_j^{(t-1)} + \eta$,
	where $\delta \beta_j^{(t-1)} = |\beta_j^{(t-1)} - \beta_j^{(t-2)}|$
	and $\eta$ is a small constant (e.g. we used $\eta=10^{-4}$).
	\item 
	\label{cond:corr}
	Group the coefficients into jobs to be dispatched
	in the $t$-th iteration, where each job contains exactly one coefficient. 
	More precisely, find a set of $P$ coefficients to be dispatched 
	such that\\[-0.4cm]
	{\footnotesize
	\begin{align}
	\{\beta_{v_{t_1}}, \ldots, \beta_{v_{t_P}}\} & \leftarrow 
	\argmin_{v_{t_1}, \ldots, v_{t_P} \in \mathcal{P}_t}{\sum_{j,k \in \{v_{t_1}, \ldots, v_{t_P}\}}{|\bx_j^T \bx_k|}} \nonumber \\
	& \mbox{ such that $|\bx_j^T \bx_k| \leq \rho$ for all $j\neq k$} \nonumber
	\end{align}
	}\\[-0.4cm]
	Here $\bx_j^T \bx_k$ represents the correlation between 
the $j$-th covariate and the $k$-th covariate; we assume $\bX$ has been standardized for Lasso. 
	\item Dispatch $\{\beta_{v_{t_1}}, \ldots, \beta_{v_{t_P}}\}$ to 
	$P$ parallel workers.
	\item Receive updated $\{\beta_{v_{t_1}}, \ldots, \beta_{v_{t_P}}\}$
	from the workers, to be used in steps 1-2 next iteration.
\end{enumerate}

Below, we present highlights from our theoretical results.
Our analysis is based on 
the sampling distribution $p(j) \propto \delta \beta_j^{(t)}$
(in practice, we approximate $\delta \beta_j^{(t)}$ with
$\delta \beta_j^{(t-1)} + \eta$ since $\delta \beta_j^{(t)}$ is unavailable at $t$-th iteration before computing $\beta_j^{(t)}$; we 
introduced $\eta$ to give all $\beta_j$s non-zero probability to account for the approximation),
and the allowed model dependency threshold $\rho$ at each iteration ---
this is unlike the global condition for all iterations used in \cite{bradley2011parallel,scherrer2012feature}.

For theoretical analysis, we rewrite problem \eqref{eq:lasso} as:
$\min_{\bbeta} L(\bX, \by, \bbeta) + \lambda \sum_{j=1}^{2J} \beta_j$,
where $\bX$ contains $2J$ features by duplicating original features with opposite sign (see appendix for details),
and $\beta_j\geq 0$, for all $j=1, \ldots, 2J$.
We define the Lasso objective as $F(\bbeta^{(t)}) = \frac{1}{2}\left\| \by - \bX \bbeta^{(t)} \right\|_2^2 + \sum_{j=1}^{2J} |\beta^{(t)}_j|$, 
and the following theorem shows that $p(j) \propto \delta \beta_j^{(t)}$ is approximately optimal for SAP.

\begin{theorem}
Suppose $\mathcal{P}_t = \{v_j\}_{j=1}^{P}$ 
is the set of indices of coefficients updated in parallel
at the $t$-th iteration, and
$\rho$ is sufficiently small such that 
$\rho \delta \beta_j^{(t)} \delta \beta_k^{(t)} < \epsilon$, 
for all $j \neq k \in \mathcal{P}_t$,
where $\epsilon$ is a small positive constant.
Then, the sampling distribution
$p(j) \propto  \frac{1}{2}(\delta \beta_j)^2$ 
approximately maximizes a lower bound $\mathcal{L}$ to
the expected decrease in the objective function $F(\bbeta^{(t)})$
after updating coefficients indexed by $\mathcal{P}_t$,
where $\mathcal{L}$ is defined as
\begin{equation}
\mathcal{L} \leq E_{\mathcal{P}_t} \left[F(\bbeta^{(t)})-F(\bbeta^{(t)}+ \Delta \bbeta^{(t)})\right]. 
\end{equation}
\end{theorem}

This means that our scheduling strategy for parallel lasso 
approximately maximizes the lower bound for the progress per iteration
(We defer the proof to the appendix).

We now discuss SAP's scalability with respect to the Shotgun algorithm, which
determines $\mathcal{P}_t$ uniformly at random~\cite{bradley2011parallel}.
Firstly, SAP always acheives the maximum effective parallelization
allowed by input data, by actively minimizing the interference caused by parallel updates.
In contrast, Shotgun's effective parallelization 
is reduced whenever the (randomly drawn) coefficients happen
to be correlated, thus producing intereference when updated in parallel.
Furthermore, SAP always chooses the $P$ coefficients
with the effort to decrease the objective function, whereas Shotgun
is agnostic to coefficient importance.
Because of these two factors, SAP has superior theoretical (and as we shall show,
empirical) scalability over Shotgun. 

\section{Experimental Results}
We show that the SAP model (implemented as \xxx{}) outperforms the unstructured model parallelism, 
which selects variables uniformly at random for parallel execution, as well as
the static block-structured parallelism model, which does not change block structures during execution.
We demonstrate this on two exemplar applications, parallel Lasso and 
parallel MF; experimental details follow:

\paragraph{Datasets}
For parallel Lasso, we used one real and one synthetic dataset. 
Our real dataset was the Alzheimer's disease (AD) dataset~\cite{sagebionetworks},
containing 463 samples and 508,999 covariates (single nucleotide polymorphisms)
for $\bX$, and real-valued \textit{APOE} gene expression levels for $\by$.
For synthetic data, we generated 450 samples with 1,000,000 features;
and a real-valued output with 10,000 true non-zero coefficients. 
For parallel MF, we used the NetFlix \cite{gemulla2011large} and Yahoo-Music \cite{yahoo-music} datasets.
The NetFlix dataset contains 480,189 users versus 17,770 movies (100,480,507 non-zero entries)
while the Yahoo-Music dataset contains 1,948,882 users versus 98,213 songs (115,579,440 non-zero entries).

\paragraph{Experimental platform and STRADS configurations}
We ran the experiments on a compute cluster, with the following machine specifications:
64 cores ($16 \times 4$ AMD Opteron 1.4 GHz), 3TB SATA drive,
128GB RAM, and 10GbE network interface.
Parallel Lasso and MF applications were tested in different platforms.
We ran the parallel Lasso application
in the distributed setting (multiple machines)
using from 60 to 240 cores,
and parallel MF in the single multi-core machine setting
using from 4 to 16 cores.
STRADS was configured as follows: for Lasso, we used 
$\eta = 10^{-6}$, $\rho = 0.1$, and $\lambda = 5 \times 10^{-4}$, 
and for MF, we partitioned variables 
such that each block contains $\left\lceil  N/P \right\rceil$ or $\left\lceil  M/P \right\rceil$ variables, where $P$ is the number of cores.

\subsection{Experiments on Parallel Lasso}
\begin{figure}[t] 
\centering
\includegraphics[width=0.8\textwidth]{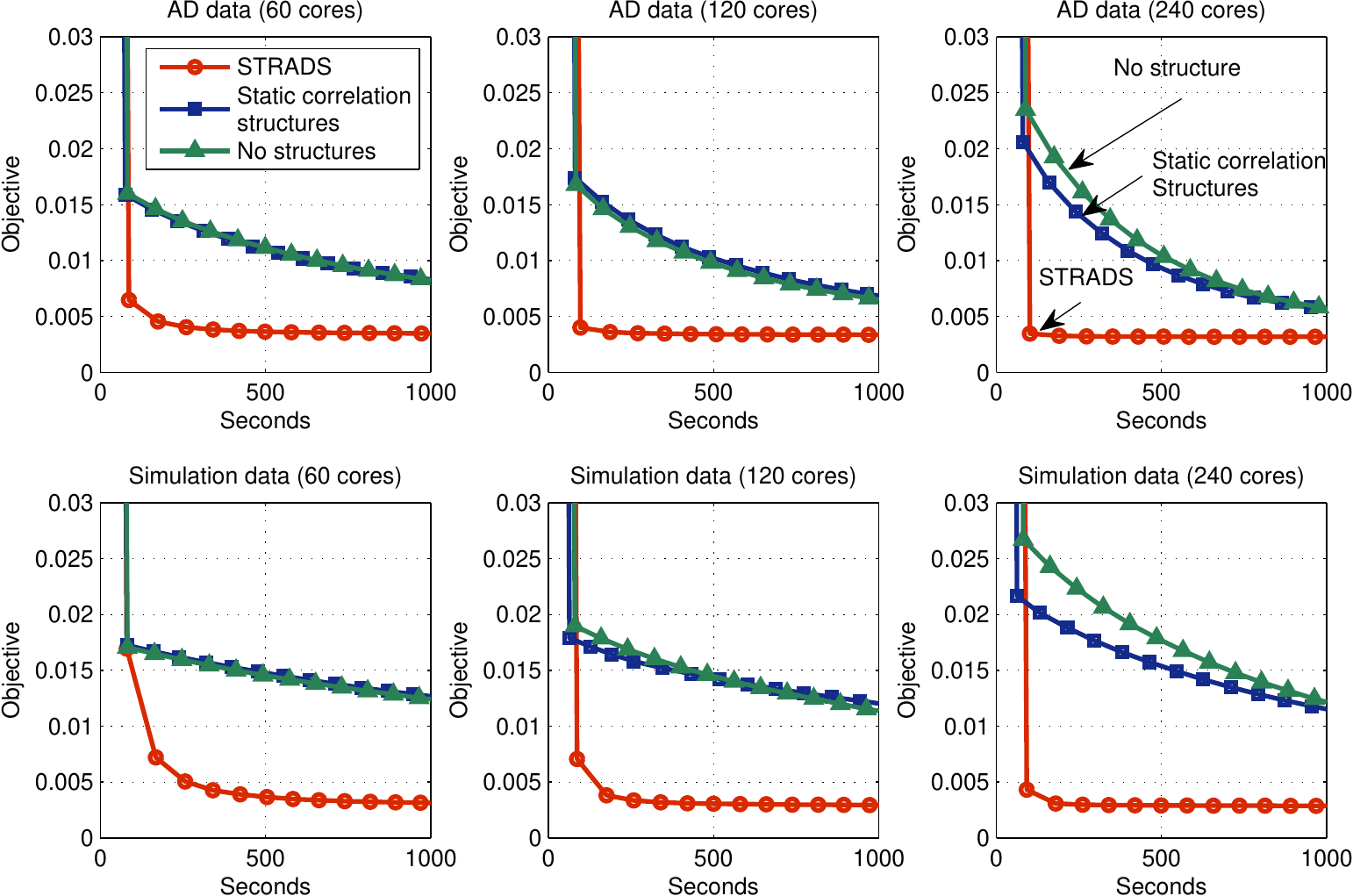}
\vspace{-0.3cm}
\caption{{\bf Distributed parallel Lasso results} for three scheduling models:
SAP (scheduled dynamic block structures using \xxx{}), static block structures,
and uniform random selection of variables (no structures).
The first row shows results for the Alzheimer's disease (AD) dataset,
while the second row corresponds to our synthetic dataset.
We vary the number of processor cores from 60 to 240.
}  
\label{fig:lasso_res}
\end{figure}

Fig. \ref{fig:lasso_res} shows objective vs. time plots for 
\xxx{} (SAP model for dynamic block structures), a static correlation scheduler
(static block structures),
and a random scheduler (no block structures), over several machine configurations. 
The static block scheduling uses the following strategy:
pick a set of variables uniformly at random, and dispatch only variables 
that are nearly independent (i.e. $< \rho$ correlation).
As for unstructured scheduling, we used the Shotgun approach \cite{bradley2011parallel},
which  selects variables uniformly at random; note that the original
Shotgun paper was limited to a single multi-core machine, whereas our experiments
bring Shotgun into the distributed setting.

The first row of Fig. \ref{fig:lasso_res} contains AD data results,
while the second row contains synthetic data results, over 60, 120, and 240 cores.
In all cases, \xxx{} converged much faster than the other two schedulers.
We point out three phenomena observed in these experiments:
first, \xxx{} consistently generates an early sharp drop in the objective function value;
this is because after all variables have been updated at least once,
\xxx{} now has a full estimate of the importance distribution $p(j)$,
so it can now prioritize more important variables.
This results in a dramatic reduction in objective value.

Second, STRADS exhibits not only a faster convergence rate, but also
a substantially better objective function value when converged. 
It is possible that the other two approaches will eventually achieve the same objective that
STRADS had. In practice however, algorithms are run with an automatic stopping condition ---
typically a minimum threshold on change in objective value. Under such a stopping condition,
STRADS achieves a better final objective value than the other schedulers.

Finally, we observe that static correlation scheduling only beats random
scheduling by a significant margin when using a large number of cores (e.g., 240).
The reason is that, with a low core count, random scheduling is unlikely
to select highly correlated variables, and hence static block structures do
not yield any benefit. Once the core count increases, the probability of picking
multiple correlated variables goes up, and static correlation scheduling
begins to show an advantage. However, \xxx{} dynamic scheduling based on
variable importance yields an even greater improvement.

\vspace{-0.2cm}
\subsection{Experiments on Parallel Matrix Factorization}
\begin{figure}[t] 
\vspace{-0.3cm}
\centering
\includegraphics[width=0.8\textwidth]{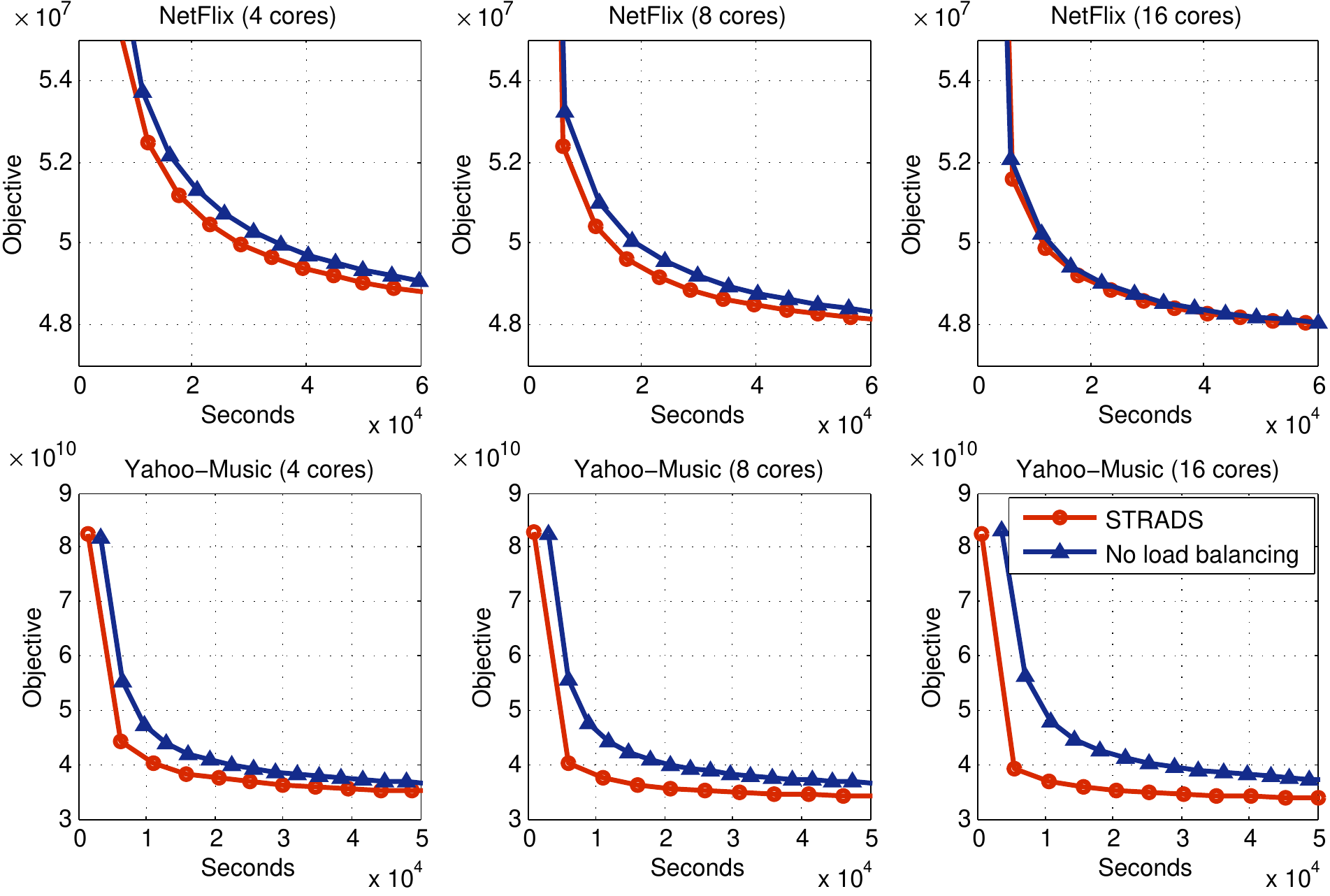}
\vspace{-0.3cm}
\caption{{\bf Single-machine parallel Matrix Factorization results}
for two scheduling models: SAP (using \xxx{}),
and a model with no load balancing.
The first row shows results for the Netflix dataset,
while the second row corresponds to the Yahoo-Music dataset.
We vary the number of processor cores from 4 to 16.}
\label{fig:mf_res}
\end{figure}

Fig. \ref{fig:mf_res} compares, for 4 to 16 cores on a single machine,
parallel MF using \xxx{}, versus a scheduler with no load balancing (that
partitions the matrix rows and columns uniformly,
without regard to the number of non-zero entries in each row/column).
This experiment is intended to demonstrate the performance gains
from load balancing through \xxx{}.

On the NetFlix dataset (first row of Fig. \ref{fig:mf_res}),
\xxx{} exhibits slightly better convergence rate 
for 4 and 8 cores, but an insubstantial benefit for 16 cores.
The reason is one of sampling statistics: when using a small number of cores/blocks and uniformly sampling over
rows and columns, the final distribution of block sizes (i.e. number of non-zero entries)
exhibits a large variance --- that is to say, some blocks can be much larger than others.
Hence, the largest block becomes a severe bottleneck. However, once the number
of cores/blocks is increased, the variance in block sizes drops, and the bottleneck is thus reduced.

For the Yahoo-Music dataset (second row of Fig. \ref{fig:mf_res}),  
\xxx{} exhibits much clearer benefits from load balancing.
Moreover, unlike the NetFlix dataset, the gain due to load balancing
actually increases with more cores. It turns out that
the non-zero entries in the Yahoo-Music dataset are heavily biased
towards a few items (i.e. strong power-law behavior) --- hence without load
balancing, algorithm performance is no better than a single thread due to bottlenecking
on the extreme users. \xxx{} load balancing resolves this problem,
allowing for full parallelism (which explains the widening
gap w.r.t. the naive scheduler at higher core counts).

\section{Related Work and Discussion}
Variable scheduling is a key component of many distributed platforms such as 
Pregel \cite{malewicz2010pregel}, 
MapReduce \cite{dean2008mapreduce} and GraphLab \cite{low2012distributed}.
For example, GraphLab paritions graph data  
to minimize communication and synchronization costs
between different connected nodes; furthermore, GraphLab provides various 
consistency schemes to synchronize dependent parameters or variables.
Pregel is designed to process large scale graphs, and 
schedules computations using workflow graphs.
Hadoop distributes the data to workers, in a manner that limits
communication due to map-reduce synchronization.
Our work differs from these scheduling approaches, in that we consider
not only static information embedded in the data, but also
dynamic information such as transient parameters or variables learned at runtime.

Algorithms for our two exemplar applications, parallel Lasso and MF, have been extensively studied in the literature:
examples include 
randomized block-coordinate descent \cite{richtarik2012iteration},
dual decomposition \cite{boyd2011distributed}, 
parallel stochastic gradient decent \cite{luo2012parallel,recht2011parallel},
and parallel coordinate descent 
\cite{bradley2011parallel,yu2012scalable}.
These works differ from ours in the sense that
we suggest a general-purpose dynamic scheduler to boost the performance and correctness of parallel ML algorithms,
rather than an algorithm tailored to a specific application.
In fact, we used existing algorithms for parallel Lasso and MF without any modification.
In that regard, STRADS can be combined with any new developments
in parallel Lasso or MF algorithms, so as to yield further performance improvements.

Future work includes harnessing STRADS to accelerate diverse Big Model applications.
By considering the unique ML properties of each application,
we can develop principles for analyzing intermediate variables/parameter values in the context of the data,
in order to formulate the importance distribution $p(j)$ and dependency function $d(\bx_j,\bx_k)$ necessary
for high performance model-parallelism with \xxx{}.
Furthermore, we will explore principled ways to improve the efficiency of \xxx{},
such as increasing the size of blocks to be dispatched while still tightly controlling interference effects
between model variables --- in order
to minimize communication costs between workers and scheduler and thus maximize CPU utilization.

\section*{Appendix: Proof of Theorem 1}

\paragraph{Preliminaries}

The $\ell_1$-regularized regression \cite{tibshirani1996regression}  
takes the form of an optimization program: 
\begin{align}
\min_{\bbeta} L(\hat{\bX}, \by, \bbeta) + \lambda \sum_{j} |\beta_j|
\label{eq:lasso_appendix}
\end{align}
where $\lambda$ denotes the regularization parameter, and $L(\cdot)$ is a non-negative convex loss function.
We assume that $\hat{\bX}$ and $\by$ are standardized and consider \eqref{eq:lasso_appendix} without an intercept.
For simplicity but without loss of generality, we let  
$L(\hat{\bX},\by,\bbeta) = \frac{1}{2}\left\| \by - \hat{\bX} \bbeta \right\|_2^2$.
However, it is straightforward to use other loss functions 
such as logistic-loss using the same approach shown in \cite{bradley2011parallel}.

For theoretical analysis, we rewrite problem \eqref{eq:lasso_appendix} as:
\begin{align}
\min_{\bbeta} L(\bX, \by, \bbeta) + \lambda \sum_{j=1}^{2J} \beta_j,
\label{eq:lasso_rewrite}
\end{align}
where $\bX$ contains duplicated features with opposite sign such that
$x_j^i = \hat{x}_{j}^i$, and
$x_{j+J}^i = -\hat{x}_{j}^i$ for all $j=1,\ldots,J$
and $i=1,\ldots, N$, and 
$\beta_j\geq 0$, for all $j=1, \ldots, 2J$.
Note that problem \eqref{eq:lasso_appendix} and \eqref{eq:lasso_rewrite} are equivalent
optimization problem \cite{bradley2011parallel}.
To optimize the problem \ref{eq:lasso_rewrite}, we can use parallel coordinate descent
method (Shotgun) proposed by \cite{bradley2011parallel}, and
the update rule is $\beta_j \leftarrow \beta_j+\delta \beta_j$, where $\delta \beta_j$ is given by,
\[
\delta \beta_j = \max\{-\beta_j, -\nabla(F(\bbeta))_j\},
\]
where $F(\bbeta) = \frac{1}{2}\left\| \by - \bX \bbeta \right\|_2^2 + \sum_{j=1}^{2J} |\beta_j|$.

\setcounter{theorem}{0}
\begin{theorem}
Suppose $\mathcal{P}_t = \{v_j\}_{j=1}^{P}$ 
is the set of indices of coefficients updated in parallel
at the $t$-th iteration, and
$\rho$ is sufficiently small such that 
$\rho \delta \beta_j^{(t)} \delta \beta_k^{(t)} < \epsilon$, 
for all $j \neq k \in \mathcal{P}_t$,
where $\epsilon$ is a small positive constant.
Then, the sampling distribution
$p(j) \propto  \frac{1}{2}(\delta \beta_j)^2$ 
approximately maximizes a lower bound $\mathcal{L}$ to
the expected decrease in the objective function $F(\bbeta^{(t)})$
after updating coefficients indexed by $\mathcal{P}_t$,
where $\mathcal{L}$ is defined as
\begin{equation}
\mathcal{L} \leq E_{\mathcal{P}_t} \left[F(\bbeta^{(t)})-F(\bbeta^{(t)}+ \Delta \bbeta^{(t)})\right]. 
\end{equation}
\end{theorem}

\begin{proof}
From assumption 3.1 in \cite{bradley2011parallel}, we have 
\begin{align}
F(\bbeta^{(t)})& -F(\bbeta^{(t)}+ \Delta \bbeta^{(t)})  \nonumber\\
& \geq -(\Delta \bbeta^{(t)})^T \nabla F(\bbeta^{(t)}) - \frac{1}{2}(\Delta \bbeta^{(t)})^T
\bX^T \bX(\Delta \bbeta^{(t)}) \nonumber
\end{align}
where $(\Delta \bbeta^{(t)})^T \nabla F(\bbeta^{(t)})\leq 0$.
For simple notation, let us omit the super script representing $t$-th iteration.

Suppose index of coefficient $j$ is drawn from a sample distribution $p(j)$, and 
a pair of indices $(i,j)$ is drawn from $p(i,j)$.
Taking expectaion with respect to $\mathcal{P}_t$: 
{\footnotesize
\begin{align}
&E_{\mathcal{P}_t} [F(\bbeta)-F(\bbeta+ \Delta \bbeta)] \geq - E_{\mathcal{P}_t}[\sum_{j\in \mathcal{P}_t} \delta \beta_j \nabla (F(\bbeta))_j] \nonumber \\
& \qquad - \frac{1}{2}E_{\mathcal{P}_t}\left[\sum_{i,j\in \mathcal{P}_t}\delta \beta_i (\bX^T \bX)_{i,j}\delta \beta_j\right]\\
&= - E_{\mathcal{P}_t}[\sum_{j\in \mathcal{P}_t} \delta \beta_j \nabla (F(\bbeta))_j + \frac{1}{2}(\delta \beta_j)^2] \\
& \qquad - \frac{1}{2}E_{\mathcal{P}_t}\left[\sum_{i,j\in \mathcal{P}_t, i\neq j}\delta \beta_i (\bX^T \bX)_{i,j}\delta \beta_j\right]\\
&= - P\sum_{j\in \mathcal{P}_t} p(j) \left[ \delta \beta_j \nabla (F(\bbeta))_j + \frac{1}{2}(\delta \beta_j)^2 \right]\\
& \qquad  - \frac{1}{2}P(P-1) \left[\sum_{i,j\in \mathcal{P}_t, i\neq j} p(i,j) \delta \beta_i (\bX^T \bX)_{i,j}\delta \beta_j\right] \nonumber \label{eq:rhoused}\\
&= - P\sum_{j\in \mathcal{P}_t} p(j) \left[ \delta \beta_j \nabla (F(\bbeta))_j  + \frac{1}{2}(\delta \beta_j)^2 \right]\\
& \qquad  - \frac{1}{2}P(P-1) \left[\sum_{i,j\in \mathcal{P}_t, i\neq j} p(i,j) \min\{\rho,(\bX^T \bX)_{i,j}\} \delta \beta_i \delta \beta_j  \right] \nonumber \label{eq:useassumption}\\
&\approx  P\sum_{j\in \mathcal{P}_t} p(j) \left[ -\delta \beta_j \nabla (F(\bbeta))_j  - \frac{1}{2}(\delta \beta_j)^2 \right].
\end{align}
}
In \eqref{eq:rhoused}, we used $p(i,j)=0$ if $(\bX^T \bX)_{i,j} > \rho$ 
because $\beta_i$ and $\beta_j$
cannot be updated in parallel if $|\bx_i^T \bx_j|>\rho$.
Recall that we find $P$ coefficients $\{\beta_{v_{t_1}}, \ldots, \beta_{v_{t_P}}\}$ 
to be updated in parallel by solving: 
{\footnotesize
	\begin{align}
	\{\beta_{v_{t_1}}, \ldots, \beta_{v_{t_P}}\} & \leftarrow 
	\argmin_{v_{t_1}, \ldots, v_{t_P} \in \mathcal{P}_t}{\sum_{j,k \in \{v_{t_1}, \ldots, v_{t_P}\}}{|\bx_j^T \bx_k|}} \nonumber \\
	& \mbox{ such that $|\bx_j^T \bx_k| \leq \rho$ for all $j\neq k$.} \nonumber
	\end{align}
	}
Further, in \eqref{eq:useassumption} we used our assumption that 
$\rho \delta \beta_j \delta \beta_k < \epsilon$, for all $j\neq k \in \mathcal{P}_t$
for small $\epsilon$.
Thus, from \eqref{eq:useassumption} the lower bound of 
$E_{\mathcal{P}_t} [F(\bbeta)-F(\bbeta+ \Delta \bbeta)]$
is maximized when 
$p(j) \sim \left[ -\delta \beta_j \nabla (F(\bbeta))_j  - \frac{1}{2}(\delta \beta_j)^2 \right]$.
Furthermore, $\delta \beta_j = \max\{-\beta_j, -\nabla(F(\bbeta))_j\}$.
Thus,  $-\delta \beta_j \nabla (F(\bbeta))_j  - \frac{1}{2}(\delta \beta_j)^2 
\leq \frac{1}{2}(\delta \beta_j)^2 $
because $\delta \beta_j \geq -\nabla(F(\bbeta))_j$.
Therefore, $p(j) \propto \frac{1}{2}(\delta \beta_j)^2$
gives us approximately optimal distribution to maximize the lower bound 
$\mathcal{L}$ of 
$E_{\mathcal{P}_t} [F(\bbeta)-F(\bbeta+ \Delta \bbeta)]$.
\end{proof}

\bibliographystyle{plain}
\bibliography{ref}

\end{document}